\documentclass[sigconf]{acmart}

\usepackage{xcolor}
\usepackage{soul}
\usepackage{url}
\usepackage{caption}
\usepackage{graphicx}
\usepackage{amsmath}
\usepackage{amsthm}
\usepackage{booktabs}
\usepackage{multirow}
\usepackage{listings}
\usepackage{array}

\usepackage[linesnumbered, ruled, boxed]{algorithm2e}

\newcommand{\nosection}[1]{\vspace{2pt}\noindent\textbf{#1.}}
\AtBeginDocument{%
  \providecommand\BibTeX{{%
    \normalfont B\kern-0.5em{\scshape i\kern-0.25em b}\kern-0.8em\TeX}}}

\setcopyright{acmcopyright}
\copyrightyear{2018}
\acmYear{2018}
\acmDOI{10.1145/1122445.1122456}

\acmConference[KDD '20]{KDD '20: SIGKDD}{August, 2020}{San Diego, CA}
\acmBooktitle{KDD'20}
\acmPrice{15.00}
\acmISBN{978-1-4503-XXXX-X/18/06}

\begin{document}


\title{A Comprehensive Analysis of Information Leakage in Deep Transfer Learning}

 \author{Cen Chen, Bingzhe Wu, Minghui Qiu, Li Wang, Jun Zhou}
 \email{{chencen.cc,fengyuan.wbz,raymond.wangl,jun.zhoujun}@antgroup.com}
\email{minghui.qmh@alibaba-inc.com}
 \affiliation{%
   \institution{Alibaba Group}
   \city{Hangzhou}
   \country{China}
 }




 \renewcommand{\shortauthors}{Trovato and Tobin, et al.}

\begin{abstract}
Transfer learning is widely used for transferring knowledge from a source domain to the target domain where the labeled data is scarce. Recently, deep transfer learning has achieved remarkable progress in various applications. However, the source and target datasets usually belong to two different organizations in many real-world scenarios, potential privacy issues in deep transfer learning are posed. In this study, to thoroughly analyze the potential privacy leakage in deep transfer learning, we first divide previous methods into three categories. Based on that, we demonstrate specific threats that lead to unintentional privacy leakage in each category.  Additionally, we also provide some solutions to prevent these threats. To the best of our knowledge, our study is the first to provide a thorough analysis of the information leakage issues in deep transfer learning methods and provide potential solutions to the issue. Extensive experiments on two public datasets and an industry dataset are conducted to show the privacy leakage under different deep transfer learning settings and defense solution effectiveness. 
\end{abstract}



\maketitle
\section{Introduction}
Transfer learning is a rapidly growing field of machine learning that aims to improve the learning of a data-deficient task by knowledge transfer from related data-sufficient tasks~\cite{pan2009survey,tan2018survey,DBLP:journals/corr/abs-1911-02685}. 
Witness the success of deep learning, deep transfer learning has been widely studied and demonstrated remarkable performance over various applications, such as
medical image classification~\cite{medical_image_tl}, electronic health data analysis\cite{ehr_data}, and credit modeling~\cite{finicail_data}. 

A fundamental block of deep transfer learning is deep neural network, which is vulnerable to different attacks aiming to detect sensitive information contained in the training dataset~\cite{p3sgd,shokri2017membership,ganju2018property}.
Moreover, in most of the real-world applications where deep transfer learning is used, the source and target datasets always reside in two different organizations. As a result, deep transfer learning also faces potential privacy threats, i.e, the client in the target organization
can leverage the vulnerability of deep learning models to detect sensitive information contained in the source organization. Specifically, applying deep transfer learning comes with the interaction between the source and
target domains. Thus, the data transmission between these domains may unintentionally disclose private information. 


Existing studies on analyzing privacy leakages focus on either general machine learning models~\cite{shokri2017membership,DBLP:conf/sp/NasrSH19} or in a federated learning setting where model is collaboratively trained by multiple clients by sharing and aggregating the gradients via a server\cite{melis2019exploiting,DBLP:conf/sp/NasrSH19}.
However, there no such study on  transfer learning paradigms. 
To this end, we are the first to provide a general categorization for deep transfer learning models based on the \textit{potential information leakages}.
This is not trivial since there are numerous methods
for deep transfer learning~\cite{pan2009survey,tan2018survey,DBLP:journals/corr/abs-1911-02685}.
Given the goal of privacy leakage analysis, we care more about the \textit{interaction manner} between source and target domains. 
Thus, we divide previous works into three categories, as illustrated in Figure 1: (1) \textit{model-based} paradigm where the whole model structure and parameters are shared (2) \textit{mapping-based} where the hidden features are shared (3) \textit{parameter-based} where the parameter gradients are shared. 
Based on that, the previous works can fall into the above categories or a hybrid of them. 
For example, fine-tuning based approaches obviously belong to the first category.
The prior work~\cite{deepcoral} is based on the mapping-based paradigm, since it uses the correlation alignment loss which further depends on the shared hidden features. Similarly, previous works that minimize the domain representation difference by variants of distribution divergence metrics such as maximum mean discrepancy also fall into the second category~\cite{long2015learning,long2017deep,Rozantsev}. Fully-shared and shared-private transfer learning models ~\cite{liu2017adversarial} can be regarded as parameter-based, as they both jointly train a shared network via gradient updates in a multi-task fashion, just to name a few. 

\begin{figure*}[t!]
    \centering
    \includegraphics[width=16cm]{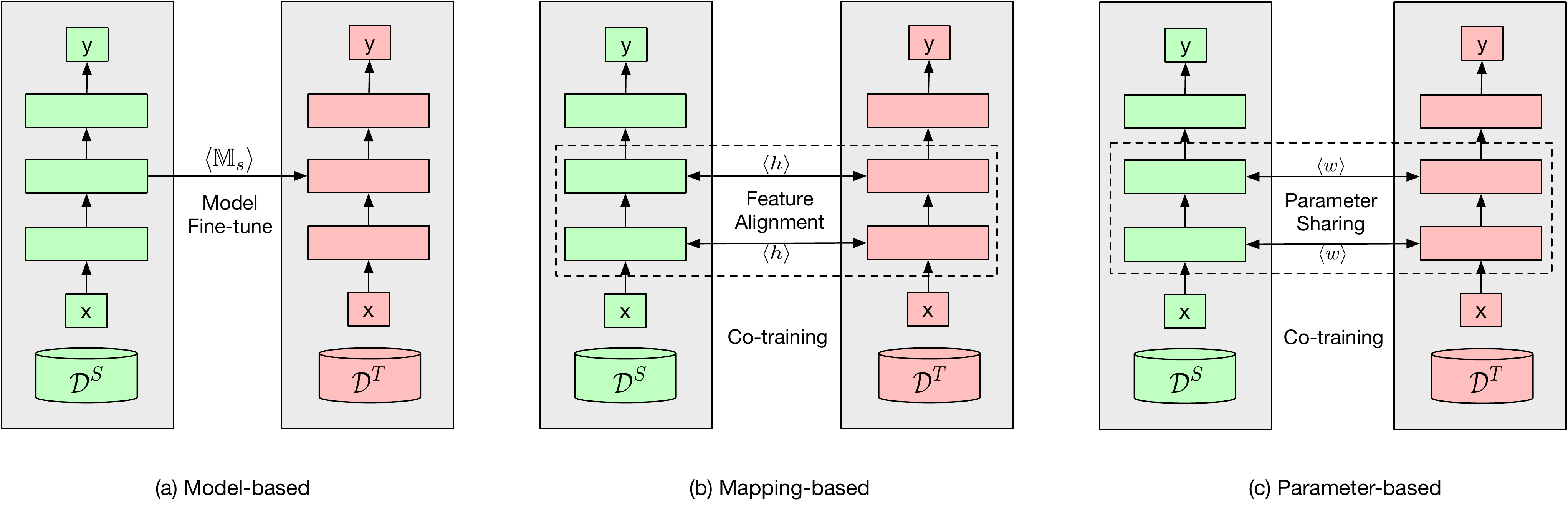}
    \caption{An illustration of model-based, mapping-based and parameter-based transfer learning paradigms. $\mathcal{D}^S$ and $\mathcal{D}^T$ denote the source domain and target domain data, respectively. Each dataset contains a set of labeled training samples that consist of the inputs $x$ and their corresponding labels $y$. We use $w$ to denote model parameters, and $h$ for feature representations.}
    \label{fig:categorization}
    \vspace{-0.1cm}
\end{figure*}

Based on the general categorization, we can build customized attacks against each paradigm and demonstrate information leakages in deep transfer learning. At a high level, we consider inferring two types of sensitive information, i.e., membership and property information.
This sensitive information can be revealed by the transmission data between the two domains as the above discussed.
Specifically, in the model-based paradigm, we build the membership attack which takes the model (learned using the source dataset) as input and determines whether a specific sample is used for training the model. In the mapping-based setting, we can build the property attack to infer properties contained in the training dataset.
For example, the attacker resides on the target domain aims to infer properties of the source domain based on the shared hidden features. In the parameter-based setting, we can similarly perform the property inference attack, i.e., the attacker can infer properties of the source domain data based on the shared gradients. More details of these attacks can be found in Section~\ref{sec:privacy_analysis}.

Empirically, to demonstrate the effectiveness of attacks, we conduct a set of experiments under the three types of transfer learning settings. Our key observation is that all these types of models do unintentionally leak information of the training data under membership/property attacks. 
Model-based paradigm is possible to leak membership information.
Parameter-based paradigm without revealing individual gradient (i.e., averaged gradients at batch-level) leaks much less property information, compared to the mapping-based paradigm where hidden features (i.e., representations at sample-level) are shared.

In summary, our main contributions are as follows:
\begin{itemize}
    \item 
    We are the first to propose a general categorization for different deep transfer learning paradigms based on their intrinsic interaction manner between the source and target domains and provide a comprehensive analysis of the potential privacy leakage profiles for each category respectively.
    \item Based on the categorization, we build specific attacks against each paradigm to demonstrate their privacy leakages.
    \item In contrast to previous works that aim to design privacy-preserving transfer learning model for a specific setting~(e.g., mapping-based ~\cite{sharma2019secure,liu2018secure}), our analysis has covered a wide range of deep transfer learning methods.
    \item We conduct extensive experiments on both public datasets and an industry marketing dataset to verify the effectiveness of our attacks and defense solutions.
\end{itemize}


\section{Preliminaries}
\label{sec:preliminary}
\subsection{Deep Transfer Learning Setting}
In this work, we focus on deep transfer learning~\cite{tan2018survey,DBLP:journals/corr/abs-1911-02685}, where the models discussed are neural network based.
Without losing generality, we consider a transfer learning setting with two domain tasks $\mathcal{T} \in\{S, T \}$, where $S$ and $T$ refer to the source domain and target domain respectively, both containing private sensitive information. 
We aim to improve the target domain learning task performance by utilizing its own data $\mathcal{D}^T$ and source domain data $\mathcal{D}^S$.
Each dataset contains a set of labeled examples $ z_i = (x_i, y_i)$, where $x$ denotes the inputs and $y$ denotes the corresponding labels.
The size of source domain data $\mathcal{D}^S$ is usually much larger than the target domain data $\mathcal{D}^T$, i.e., $N^S >> N^T$.
The goal of a domain task is to learning a transformation function $f_w ( \mathbf{x} , \mathbf{y} ) :  \mathbb{R}^d  \rightarrow \mathbb{R}$, parameterized by model weights $\mathbf{w}$. 


\subsection{Inference Attacks for DNN Models} 
The basic idea of the inference attack is to exploit the leakages when a model is being trained or released to reveal some unintended information from the training data.  
In this section, we briefly present two types of inference attacks, i.e., membership and property attacks, for machine learning models. 




\nosection{Membership Inference Attack}
Membership inference is a typical attack that aims to determine whether a sample is used as part of the training dataset. 
Membership inference may reveal sensitive information that leads to privacy breach. For example, if we can infer a patient's existence in the training dataset for a medical study of a certain disease, we can probably claim that this patient has such a disease. 
Recent works have demonstrated the membership attack attempts for machine learning models under the black-box setting \cite{shokri2017membership,long2017towards,salem2018ml}, white-box setting \cite{melis2019exploiting,nasr2019comprehensive} or both \cite{hayes2019logan}.
Shadow training is a widely adopted technique for membership inference, where multiple shadow models are trained to mimic the behavior of the target model\cite{shokri2017membership,long2017towards,salem2018ml}.
This technique assumes the attacker to have some prior knowledge about the population from the targeted model training dataset was drawn.
Recent works by \cite{melis2019exploiting,nasr2019comprehensive} explicitly exploit the vulnerabilities in gradient updates to perform attacks with white-box access. 

\begin{table*}[t!]
\caption{General categorization of different deep transfer learning paradigms and their respective leakage profiles. Without loss of generality, we assume a semi-honest threat model with attacker to be the target domain dataset owner. Three types of transfer learning paradigms require different information interactions and training strategies, thus pose different potential leakage profiles and may suffer from different inference attacks.}
\begin{tabular}{l p{9.2cm} l c c}
\hline
\multirow{2}{*}{Categorization} & \multirow{2}{*}{Brief Description} & Training  & Leakage & Inference  \\ 
 & & Strategy & Profile & Attack Type \\ \hline
Model-based & Model fine-tuning, i.e., continue training on the target domain. 
& Self-training & $\left \langle \mathbb{M}^S \right \rangle$ &  Membership\\
Mapping-based & Hidden representations alignment, i.e., reducing distribution divergence. 
& Co-training & $\left \langle h^S, h^T \right \rangle$  &  Property \\
Parameter-based & Jointly updating shared partial network, i.e., hard parameter sharing. 
& Co-training & $\left \langle w^c  \right \rangle$  &  Batch property\\
\hline
\end{tabular}
\label{tab:categorization}
\vspace{-0.1cm}
\end{table*}

\nosection{Property Inference Attack}
Another common type of attack is property inference that 
aims to reveal certain unintended or sensitive properties (e.g., the fraction of the data belongs to a certain minority group) of the participating training datasets that the model producer does not intend to share when the model is released. 
A property is usually uncorrelated or loosely correlated with the main training task.
Pioneer works of \cite{ateniese2015hacking,fredrikson2015model,ganju2018property}
conducted the property attacks that characterize the entire training dataset. While, \cite{melis2019exploiting} aimed to infer properties for a subset of the training inputs, i.e., in terms of single batches which they termed as single-batch properties. 
In this regard, membership attack can be viewed as a special case of property attack when scope for property attack is on a sample basis.


The above-mentioned two types of attacks are closely related.
Most of existing works perform those attacks against general machine learning models, while a few focus on the federated learning and collaborative learning scenarios \cite{nasr2019comprehensive,melis2019exploiting}. None of the studies systematically explore the inference attacks explicitly for the context of deep transfer learning.

\section{Privacy Leakage in Deep Transfer Learning}
\label{sec:privacy_analysis}

In this section, we first provide a general categorization of deep transfer learning according to the interaction manner between source and target domains.
Then, based on the categorization, we conduct privacy analysis through building specific attacks against different transfer learning paradigms. 


\subsection{General Categorization of Deep Transfer Learning}
Different types of transfer learning models have been proposed over the years, depending on how the knowledge is shared. 
Although in transfer learning, there may exist several existing categorizations in the literature~\cite{pan2009survey,tan2018survey,DBLP:conf/sp/NasrSH19}, 
\textit{we categorize the different deep transfer learning models based on how the two domains interact, their training strategy (e.g., co-training or self-training), and the potential leakages}. 
Broadly speaking, those transfer learning models can be categorized into three types, i.e., \textit{model-based}, \textit{mapping-based}, and \textit{parameter-based}, as illustrated in Figure~\ref{fig:categorization}. 
\begin{itemize}
    \item Model-based (in Figure 1(a)) is a simple but effective transfer learning paradigm, where the pre-trained source domain model is used as the initialization for continued training on the target domain data. Model-based fine-tuning has been broadly used to reduce the number of labeled data needed for learning new tasks/tasks of new domains~\cite{he2016deep,howard2018universal,devlin2019bert}. 
    
    \item Mapping-based (in Figure 1(b)) methods aim to align the hidden representations by explicitly reducing the marginal or conditional distribution divergence between source and target domains. 
    More specifically, alignment losses, i.e., usually in forms of distribution discrepancies/distance metrics such as MMD and JMMD, between domains are measured by such feature mapping approaches and minimized in the loss functions~\cite{long2015learning,long2017deep,Rozantsev}.
    
    \item Parameter-based (in Figure 1(c)) methods transfer knowledge by 
    jointly updating a shared partial network to learn transferable feature representations across domains in a multi-task fashion. This type of methods is mainly achieved by parameter sharing. Regardless of the design differences, they all utilize a shared network structure to transform the input data into domain-invariant hidden representations~\cite{NIPS2014_5347,liu2017adversarial,yang2017transfer}.
\end{itemize}
Based on the general categorization, we will discuss the threat model, information leakage and the customized attacks against each transfer learning paradigm in detail.

\subsection{Threat Model}
In this work, we assume all the parties 
, i.e., the domain-specific data owners, are \textit{semi-honest},  where they follow exactly the computation protocols but may try to infer as much information as possible when interacting with the other parties. 
More specifically, we work on the threat model under a deep transfer learning setting with two domains.
Without loss of generality, we assume the owner of the target dataset to be the \textit{attacker}, who intentionally attempts to infer additional information from source domain data beyond what is explicitly revealed. 
Depending on the different deep transfer learning categorizations, the attacker may have different access to the source domain information.
Note that we can naturally extend to the case where the attacker is the owner of the source dataset or the transfer learning setting with more than two data sources, however, it is not the focus of this paper.

\subsection{Privacy Analysis}
As presented in Table \ref{tab:categorization}, three types of model categorization require different information interactions and training strategies, 
thus posing different potential leakage profiles. More specifically:

\begin{itemize}
    \item Model-based methods rely on the pre-trained source domain model solely. Despite the necessity for a pre-trained model to be disclosed to the target domain, both the source and target training processes can be entirely separately, thus the potential leakage is only the final source domain model $\left \langle \mathbb{M}^S\right \rangle$.
    In this case, the attacker has full access to the source domain model including both the model structure and parameters.
    
    \item Mapping-based methods are optimized in a co-training fashion and hidden presentations of both domains have to interact with each other to measure the alignment losses or domain regularizers. We denote 
    $h_{ik}^S$ and  $h_{ik}^T$ as the hidden representation of layer $i$ at a training iteration $k\in [1, 2,..., K]$ for source and target domains, respectively.
    Specifically, such a feature matching process demands the hidden presentations, i.e., $h_{ik}^S$ and  $h_{ik}^T$, of both domains to be exposed and aligned to reduce the marginal or conditional distribution divergence between domains.
    As a result, $\left \langle h^S, h^T \right \rangle$ can potentially leak information from the training data. 
    
    \item Parameter-based methods jointly updating the shared partial network. 
    The interactions between the source and target happen when exchanging the gradients to sync the shared network parameters. 
    Let $w_k^c$ denote the model parameters for the shared network structure. At each training iteration  $k$,  $w_k^c$ is updated by averaging the gradients $\bigtriangledown_{W}$ learned by a mini-batch of training examples $z_i$ sampled from a domain dataset (either source or target). In such an alternating process, $w_k^c$ has to be revealed across domains at each iteration $k\in [1, 2,..., K]$. Thus, the potential leakage profile $\left \langle w_k^c \right \rangle$ contains all the intermediate $w_k^c$ during the training process. 
    
\end{itemize}


As a summary, in this paper, we focus on inferring information that is unintentionally disclosed by the data transmission between the source and target domains, such as membership information of an individual sample and property information of a specific sample or a subset of samples (see more details in the next subsection.). 

\subsection{Inference Attacks for Deep Transfer Learning}
Inference attacks against deep learning models generally exploit implicit memorization in the models to recover sensitive information from the training data.
Previously studies have proven that information can be inferred from the leakage  
profile, such as \textit{membership information} that decides whether a sample is used for training \cite{li2013membership,shokri2017membership,long2018understanding,melis2019exploiting,wu_sgld},
\textit{properties} which reveals certain data characteristics, 
or reconstructed \textit{feature values} of the private training records. 

In this part, to empirically evaluate the privacy leakage in different transfer learning settings (shown in Figure~\ref{fig:categorization}), we present concrete attack methods for each setting.
Note, we assume the \textit{attacker to be the owner of the target dataset} for all the three transfer learning paradigms discussed above. 

\nosection{Model-based}
As illustrated in Table \ref{tab:categorization}, the only leakage source in this setting is the model trained using data from the source domain. 
For simplicity, we denote the trained model as 
$\mathbb{M}^S :{y}=\mathbf{f}_{w}^{S}\!\left(\mathbf{x}\right)$, where $y$ is the prediction label. 
According to the training protocol in the model-based setting, the attacker can obtain the white-box access of $\mathbb{M}^S$, i.e., its structure and parameters. Thus, based on recent works~\cite{shokri2017membership,long2017towards,salem2018ml,melis2019exploiting,nasr2019comprehensive}, an attacker can design powerful membership attack methods to detect sensitive information contained in the source domain (i.e., membership attack). 
In the context of membership attack, the goal of the attacker is to train a \emph{membership predictor}, which can take a given sample $x$ as input and output the probability distribution that indicates whether the given sample is used for training the source domain model $\mathbb{M}^S$. Formally, we denote the membership predictor as $\mathcal{A}_{mem}: x, \mathbb{M}^S \to \{P(m=1|x), P(m=0|x)\}$. Here, $m=1$ means the given input $x$ is used for training the model $\mathbb{M}^S$. 

In this paper, we employ a widely used technique named \emph{shadow model training} for building the membership predictor. Specifically, we assume the attacker can have extra knowledge about the source dataset $\mathcal{D}^S$. 
The extra prior knowledge in our setting is the shadow training dataset $\mathcal{D}_{shadow}$ that comes from the same underlying distribution as $\mathcal{D}^S$ for training the source model.
The core idea of the shadow training dataset is to first train multiple shadow models that mimic the behavior of the source model. Then, the attacker can extract useful features from the shadow models/datasets and build a machine learning model that characterizes the relationship between the extracted features and the membership information.

To be specific, the attacker first evenly divides the shadow training dataset $\mathcal{D}_{shadow}$ into two disjoint datasets $\mathcal{D}_{shadow}^{train}$ and $\mathcal{D}_{shadow}^{out}$. Then the attacker trains the shadow model that has the same architecture as the source model using $\mathcal{D}_{shadow}$. 
Subsequently, features of samples from both $\mathcal{D}_{shadow}^{train}$ and $\mathcal{D}_{shadow}^{out}$ can be extracted. For each sample in $\mathcal{D}_{shadow}$, the attacker can use the output prediction vector of the shadow model as the feature following the prior work~\cite{shokri2017membership}. And each feature vector is labeled with 1 (member, if the sample is in $\mathcal{D}_{shadow}^{train}$) or 0 (non-member, if the sample
is in $\mathcal{D}_{shadow}^{out}$). At last, all the feature-label pairs are used for training the membership predictor.

Once the membership predictor is obtained, we can use it to predict the membership label of a sample in $\mathcal{D}^S$, i.e., feeding the output vector of the source domain model to the predictor.

\nosection{Mapping-based} 
In the mapping-based setting, information leakage can be both posed at the source and target domains since the training protocol proceeds in an interactive manner.  
To keep the analysis consistency, we refer the attacker to be the owner of the target domain dataset. 

The core idea of the mapping-based method is to align the hidden features extracted from the source and target domains, i.e., reducing the discrepancy between feature distributions of these two domains. This can be done by minimizing some pre-defined alignment loss function (e.g., maximum mean discrepancy\cite{long2015learning,long2017deep,Rozantsev,DBLP:conf/sp/NasrSH19}). Thus, this method will lead to the hidden features of the source domain share a similar distribution to the features of the target domain, which comes with a strong privacy implication, i.e., the attacker can leverage the feature similarity to build the attack model to detect sensitive information contained in the source domain.

We consider the property attack~\cite{ganju2018property} in this setting. At a high level, the property attack aims to infer whether a feature comes from the source domain has a specific property or not. Here, we take the case of the $k$-th iteration as an example. Given a specific property, the attacker can first collect an auxiliary dataset for assisting the attack.
Specifically, the attacker divides the target domain dataset $\mathcal{D}^T$ into two subsets, namely, $\mathcal{D}^T_{prop}$ which contains samples with the property and $\mathcal{D}^T_{nonprop}$ which consists of samples without the property. 
Subsequently,  $\mathcal{D}^T_{prop}$ and $\mathcal{D}^T_{nonprop}$ are used as the auxiliary datasets for building the attack model. 
At the $k$-th iteration, given the current parameter $w_k$ of the model trained using the target dataset, the attacker calculates the hidden features $h_k^T$ of the alignment layer $k$ with respect to the samples in the auxiliary dataset. Hidden features are labeled with 1 if the corresponding sample is in $\mathcal{D}^T_{prop}$ and 0 if the sample is in $\mathcal{D}^T_{nonprop}$. Once the attacker collects
all the feature-label pairs, she can train a property predictor $\mathcal{A}_{prop}$ using these pairs. The whole procedure is demonstrated in Algorithm~\ref{alg:mapping-based attack}.

\begin{algorithm}[t]
\DontPrintSemicolon
\caption{Property Attack}\label{alg:mapping-based attack}
\textbf{Inputs:} \\
\quad Model parameters $w_k$(on target domain)\\
\quad Auxiliary datasets: $\mathcal{D}^{T}_{prop}$ and $\mathcal{D}^{T}_{nonprop}$\\
\textbf{Outputs:}\\
\quad Property Predictor $\mathcal{A}_{prop}$\\
$P_{prop} \leftarrow []$\\
$P_{nonprop} \leftarrow []$\\
\For{each sample {\rm s} in $\mathcal{D}^{T}_{prop}$}
{
 Calculate the hidden feature $h_k^{T}$\\
 Append $\{h_k^{T}, 1\}$ to $P_{prop}$\\
}
\For{each sample {\rm s} in $\mathcal{D}^{T}_{nonprop}$}
{
 Calculate the hidden feature $h_k^{T}$ of {\rm s}\\
 Append $\{h_k^{T}, 0\}$ to $P_{nonprop}$\\
}
Train $\mathcal{A}_{prop}$ using $P_{prop}\cup P_{nonprop}$\\
\textbf{return} $\mathcal{A}_{prop}$\\
\end{algorithm}

Based on the property predictor obtained above, the attacker can conduct an online attack. Specifically, in the joint training process, at the beginning of the $k$-th iteration, the attacker receives a batch of hidden features $h^{S}_k$ from the source domain. Then, the attacker can employ $\mathcal{A}_{prop}$ to predict the property information contained in the source domain dataset.

\nosection{Parameter-based} In the parameter-based setting, information leakage is posed by the weight parameter interaction between the source and target domains. Similar to previous settings, we refer the attacker to be the owner of the dataset of the target domain. 

We consider the batch property attack in this setting. The intuition behind
this attack is that the attacker can observe the updates of the share layers calculated using a mini-batch of samples from the source domain. Thus the attacker can train a batch property predictor $\mathcal{A}_{bprop}$ to infer whether the update based on the mini-batch with a given property or not.
We take the $k$-th iteration as an example to demonstrate how the attacker conducts the attack. 
We assume that the attacker has auxiliary dataset $D^{aux}$ consisting of samples from the distribution that is similar to the source domain distribution. 
Note that, this assumption is commonly used in various previous works~\cite{sharma2019secure,collabrative_leakge}. Given a specific property, the attacker can further divide $D^{aux}$ into two sub-datasets, namely, the dataset which has the property ($D^{aux}_{prop}$) and the dataset without the property ($D^{aux}_{nonprop}$).

Based on the above setting, at the $k$-th iteration, the attacker can receive the fresh parameter $w_k$ of the shared layer which is updated based on samples from the source dataset. Then, the attacker can calculate gradients $g_{prop}$ using the mini-batch sampled from $D^{aux}_{prop}$ and $g_{nonprop}$ based on samples from $D^{aux}_{nonprop}$. 
The batch gradients based on $D^{aux}_{prop}$ are labeled with 1 and others are labeled with 0. Based on these gradient-label pairs, we can train the batch property predictor $\mathcal{A}_{bprop}$. The whole procedure is shown in Algorithm~\ref{alg:parameter-based}.

Once $\mathcal{A}_{bprop}$ is obtained, the attacker can use it to predict the btach property information of the source domain. Specifically, the attacker first generates the gradient $g_{online}$ based on the  parameters of the current and last iteration (i.e., $w_k$ and $w_{k-1}$). Then she can feed $g_{online}$ to $\mathcal{A}_{bprop}$ to obtain the prediction result. 

\begin{algorithm}[t]
\DontPrintSemicolon
\caption{Batch Property Attack}\label{alg:parameter-based}
\textbf{Inputs:} \\
\quad Model parameters $w_k$ of the shared partial network\\
\quad Auxiliary datasets: $\mathcal{D}^{aux}_{prop}$ and $\mathcal{D}^{aux}_{nonprop}$\\
\quad Sample size: $L_{prop}$, $L_{nonprop}$\\
\textbf{Outputs:}\\
\quad Batch Property Predictor $\mathcal{A}_{bprop}$\\
$P_{prop} \leftarrow []$\\
$P_{nonprop} \leftarrow []$\\
\For{l=1 to $L_{prop}$}
{
Sample mini-batch $b_{prop}$ from $\mathcal{D}^{aux}_{prop}$\\
 Calculate the gradient $g_{prop}$\\
 Append $\{g_{prop}, 1\}$ to $P_{prop}$\\
}
\For{l=1 to $L_{nonprop}$}
{
Sample mini-batch $b_{nonprop}$ from $\mathcal{D}^{aux}_{nonprop}$\\
  Calculate the gradient $g_{nonprop}$\\
 Append $\{g_{nonprop}, 0\}$ to $P_{nonprop}$\\
}
Train $\mathcal{A}_{bprop}$ using $P_{prop}\cup P_{nonprop}$\\
\textbf{return} $\mathcal{A}_{bprop}$\\
\end{algorithm}

\section{Experiments}
\label{sec:exp}
In this section, we empirically study the potential information leakages 
in deep transfer learning. We start by describing the datasets and the experimental setup. We then discuss in detail the results for various inference attacks under the aforementioned three transfer learning settings, followed by the examination of viable defenses.

\subsection{Datasets and Experimental Setup}
We conduct experiments on two public and one industry datasets, i.e., Review, UCI-Adult, and Marketing. The Review dataset is used to examine membership attack as fine-tuning methods are often used in such NLP task and typically suffer from membership inference. While the UCI-Adult dataset has some sensitive properties, thus is used for conducting property based attacks. 
Besides, the Marketing dataset is sampled from real-world marketing campaigns to examine the defense solution. Data statistics are in Table \ref{tab:data}. 

\nosection{Review} 
We construct our dataset from the public available Amazon review data \cite{DBLP:conf/recsys/McAuleyL13}, with two categories of products selected, i.e., `Watches' and `Electronics.
In our transfer learning setting, the data-abundant domain `Electronics' is viewed as the source domain, while 
the data-insufficient domain `Watches' is treated as the target. 
Each sample consists of a quality score and a review text.

Following the literature on this task
\cite{DBLP:conf/www/ChenQYZHLB19}, we adopt the TextCNN \cite{Kim14} as the base model for textual representation in the transfer learning model.
For TextCNN, we set the filter windows as 2,3,4,5 with 128 feature maps each. The max sequence length is set as 60 for the task. 
We initialize the embedding lookup table with pre-train word embedding from GloVe with embedding dimension set as 300. 
The batch sizes for training the source domain model, the shadow model, and the attack model are set as 64.

\nosection{UCI-Adult}
We consider a second Adult Census Income dataset~\cite{kdd/Kohavi96} from the UC Irvine repository to examine the property attack issues. The dataset has 14 properties such as country, age, work-class, education, etc. To form the transfer learning datasets, we use data instances with country of ``U.S.'' as source domain and ``non-U.S.'' as the target. 
For the UCI dataset, we consider an MLP as our base model for the transfer learning models.

\nosection{Marketing}
For transfer learning purpose, two sets of data are sampled from two real-world marketing campaigns data that contains user profile, behaviour features and adoptions. One data-abundant campaign is used as source domain, while the other is used as the target. The task is to predict whether a user will adopt the coupon.


\begin{table}[t!]
\caption{Data statistics and tasks used for experiment.}
\vspace{-0.1cm}
\begin{tabular}{l l r}
\hline
Dataset & Statistics & Task \\ \hline
\multirow{2}{*}{Amazon Review} 
& Electronics:  \#354,301 & Review Quality\\
& Watches:       \#9,737 &  Prediction\\\hline
\multirow{4}{*}{UCI Adult}
& Source Train: \#29,170 & \\
& Source Test: \#14,662 & Census Income\\
& Target Train: \#3391 & Prediction\\
& Target Test: \#1619 & \\\hline
\multirow{2}{*}{Marketing}
& Source:  \#236,542 & Coupon Adoption \\
& Target:  \#140,964 &  Prediction\\\hline
\end{tabular}
\label{tab:data}
\vspace{-0.2cm}
\end{table}

\nosection{Implementation Details}
Specifically, for model-based setting, a fine-tuning approach is adopted where both the source domain model and the shadow model employed the same model structure, i.e., the above-mentioned TextCNN structure, followed by an MLP layer of size 128. 
The attack model adopted is an MLP with network of [16, 8].
For the mapping-based setting, we consider an MLP with a network structure of [64, 8] for source and target models, an MMD metric is used as the alignment loss between the two domains, and another MLP with network of [64, 8] for the attack model. 
For parameter-based setting,  we consider a fully-shared model structure with an MLP as the base model with structure of [64, 8] at the task training stage, and another MLP with network of [16, 8] for the attack model.

All the above neural network based transfer learning models are implemented with TensorFlow 
and trained with NVIDIA Tesla P100 GPU using Adam optimizer, if not specified. The learning rate is set as 0.001. The activation function is set as ReLU. 

For all the settings, we use Area Under Curve (AUC) to evaluate the overall attack performance. 
For membership attack, we also adopt precision for evaluation, as membership of the training dataset is the concern. Precision is defined as the fraction of samples predicted as members are indeed members of the training data. 
For property attacks, we further include accuracy for evaluating the overall attack performance.

\subsection{Model-based Setting}
In the model-based setting, the attacker has the full white-box access to the source model, thus the attacker is able to train shadow models to mimic the behavior of the source domain mode. In this setting, we explore the possibility of performing membership inference on the source model. Despite the context differences, membership attacks for model-based transfer learning models can be conducted the same way as for any trained stand-alone deep learning models\cite{shokri2017membership,salem2018ml,collabrative_leakge}.
\begin{table}[t!]
\caption{Membership attack for model-base settings.}
\vspace{-0.1cm}
\label{tab:model-based}
\begin{tabular}{l | l l | l l}
\toprule
 & Train Acc & Test Acc & Attack AUC & Attack Prec. \\
\hline
10 epoch & 0.9290 & 0.7182 & 0.5014 & 0.4705 \\
20 epoch & 0.9338 & 0.8648 & 0.7355 & 0.5868 \\
30 epoch & 0.9589 & 0.8563 & 0.6744 & 0.5832  \\
\bottomrule
\end{tabular}
\vspace{-0.1cm}
\end{table}

To build the membership predictor introduced in Section~\ref{sec:privacy_analysis}, we split the original source dataset into three parts, namely, training dataset (248,011 samples), test dataset (70,860 samples), and shadow dataset (70,860 samples). We set the number of shadow models to 3. 
Table~\ref{tab:model-based} shows the training and testing accuracy for the attack classifier on the shadow dataset and the attack performance, i.e., AUC and precision, on the source domain model. 
Smaller gaps between training and testing accuracy suggests better generalization and predictive power of the attack classifier. Overall, we find that even in the model-based setting without direct interactions between the source and target domains during the training process, the source domain model does leak a considerable amount of membership information.
We also examine the effect of over-fitting by adjusting the number of epoch trained for both the source domain and shadow models.
We observe the more over-fitted a model, the more information can be potentially leaked. However, it decreases when the number of epochs furthers increases. This echos the findings in~\cite{shokri2017membership} that over-fitting is an important factor but not the only factor that contributes to the leakage of membership information.

\subsection{Mapping-based Setting}
We then investigate property attacks under the mapping-based transfer learning setting
to infer the properties of the source domain data during the training process. The properties are not the same as the prediction label, or even independent of the prediction task. To examine this, we construct two more datasets based on the UCI adult dataset: Prop-race and Prop-sex, discussed as follows.
\begin{itemize}
    \item Prop-sex: we filter out the property ``sex'' feature from the input features in UCI-Adult data and use it as the attack task label. The attack label is set as 1 if the property ``sex'' is male and 0 otherwise. This is to examine whether a attacker can infer training instance's gender 
    during the training process.
    \item Prop-race: similarly we filter out the ``race'' feature to study whether this property appears in training instances. Attack label is set as 1 if property ``race'' is white and 0 otherwise. 
\end{itemize}

Table~\ref{tab:mapping-based} shows the results of the property attack on the Prop-sex dataset. The attack has a good AUC of 0.77 in the property inference task, which shows the transfer learning setting does have information leakage. The attack precision is high on the property ``sex:male'', and less satisfactory on ``sex:female''. 
The female property has much less instances and thus may be harder to be predicted.

We have also conducted experiments on another property ``race'' in Table~\ref{tab:mapping-based-race}, the attack AUC is lower than on Prop-sex, 0.5885 vs. 0.7766. These results demonstrate that Prop-sex leaks more property information than Prop-race in terms of attack AUC.  At the first glance, ``race: white'' has much higher attack precision than ``race:non-white'', 0.8901 vs. 0.2179. Close examination shows  ``race: white'' dominates the training data with around 80\% label coverage.

\begin{table}[]
\caption{Property attack on ``sex'' for mapping-base settings.}
\vspace{-0.1cm}
\label{tab:mapping-based}
\begin{tabular}{l | l l | c | c c}
\toprule
\multirow{2}{*}{Prop-sex} & \multicolumn{2}{c|}{Test} & Attack & \multicolumn{2}{c}{Attack Prec.} \\
& AUC & Acc & AUC & male & female \\
\hline
2 epoch & 0.8513 & 0.8011 & 0.7357 & 0.7010 & 0.5918 \\
5 epoch & 0.8522 & 0.8101 & 0.7435 & 0.7072 & 0.6077 \\
10 epoch & 0.8650 & 0.8171 & 0.7766 & 0.7132 & 0.5883 \\
\bottomrule
\end{tabular}
\end{table}

\begin{table}[]
\caption{Property attack on ``race'' for mapping-base settings.}
\vspace{-0.1cm}
\label{tab:mapping-based-race}
\begin{tabular}{l | l l | c | l l}
\toprule
\multirow{2}{*}{Prop-race} & \multicolumn{2}{c|}{Test} & Attack & \multicolumn{2}{c}{Attack Prec.} \\
& AUC & Acc & AUC & white & non-white \\
\hline
2 epoch & 0.7640 & 0.8000 & 0.5661 & 0.8812 & 0.1652 \\
5 epoch & 0.7680 & 0.8012 & 0.5815 & 0.8914 & 0.1845 \\
10 epoch & 0.7750 & 0.8010 & 0.5885 & 0.8901 & 0.2179 \\
\bottomrule
\end{tabular}
\end{table}

We further examine the correlation between the chosen properties and the underlying Census Income prediction task, and compare it with the property attack results in Table~\ref{tab:property-corr}. In general, both Prop-race and Prop-sex have information leakage issues, and the property with a higher negative/positive correlation with the main prediction task tends to cause higher potential information leakage.

\begin{table}[]
\caption{The Pearson correlation between the main prediction task and the property label vs. property attack results.}
\label{tab:property-corr}
\vspace{-0.1cm}
\begin{tabular}{l l l l}
\toprule
 & Correlation & Attack AUC & Attack Acc\\
\hline
Prop-race & -0.0837 & 0.5885 & 0.5646\\
Prop-sex & -0.2146 & 0.7766 & 0.7059\\
\bottomrule
\end{tabular}
\vspace{-0.2cm}
\end{table}

\subsection{Parameter-based Setting}
For the parameter-based setting, we further process the UCI adult data to perform the batch property attack. Again we use these two properties ``race'' and ``sex'' and form two datasets namely BProp-race and BProp-sex respectively. For both datasets, we set the batch size as 8. For each batch in BProp-race, we set the label as 1 if a batch has at least one data instance containing the ``non-white'' property and 0 otherwise. For BProp-sex, we set the label as 1 if a batch has at least one instance containing the ``female'' property and 0 otherwise. The data statistics are shown in Table~\ref{tab:net-based-data}. We find for BProp-race, the positive instance ratio in the source domain is drastically different from the target, this may bring negative transfer. While for the BProp-sex, the positive instance ratio is similar for both domains.

As shown in Algorithm~\ref{alg:parameter-based}, the selection of $\mathcal{D}^{aux}$ is the key to building the batch property predictor. In this paper, we directly use the target dataset as the auxiliary dataset and find it works in practice, as both domains commonly are related for the transferring purpose. Generally, we can obtain better attack performance if we can use part of source domain samples to form the auxiliary dataset.
The results are presented in Table~\ref{tab:parameter-based}. First, we find the attack AUC for the batch property attack in the parameter-based setting are generally not as high as those from the property attack in the mapping-based setting. The AUC is around 0.56 in batch property attack, while up to 0.77 for property attack. This shows the parameter-based setting is generally less vulnerable from attacks, which can possibly be justified by the fact that the gradients exchanged have been averaged at the batch-level.
Second, the results for BProp-sex are better than BProp-race, as AUC results are 0.5654 and 0.5545 respectively. 
This is intuitive as observed from Table~\ref{tab:net-based-data}, the domain difference in BProp-race is larger than the BProp-sex. The model performance in BProp-race may suffer from the negative transfer learning due to the domain gap, thus may lead to the decrease in attack performance. Overall, the parameter-based method is possible to suffer from batch property attacks, however, the information leakage problem is not that severe.

\begin{table}[]
\caption{Batch property attack data statistics.}
\vspace{-0.1cm}
\label{tab:net-based-data}
\begin{tabular}{l | c c|  c c}
\toprule
& \multicolumn{2}{c|}{Source data size} & \multicolumn{2}{c}{Target data size} \\
& Positive & Negative & Positive & Negative \\\hline
BProp-race & 38 & 528 & 2225 & 2637 \\
BProp-sex & 65 & 501 & 418 & 4444 \\
\bottomrule
\end{tabular}
\end{table}

\begin{table}[]
\caption{Batch property attack for parameter-based settings.}
\vspace{-0.1cm}
\label{tab:parameter-based}
\begin{tabular}{l | c c | c c}
\toprule
& \multicolumn{2}{c|}{Test} & \multicolumn{2}{c}{Attack} \\
& AUC & Acc & AUC & Acc \\
\hline
BProp-race & 0.8466 & 0.8054 & 0.5545 & 0.5372 \\
BProp-sex & 0.8577 & 0.8331 & 0.5654 & 0.9140 \\
\bottomrule
\end{tabular}
\vspace{-0.2cm}
\end{table}


\subsection{Defense Solutions}
A standard way to prevent statistical information leakage is differential privacy. However, the privacy guarantee provided by differential privacy comes with the decreasing of the model utility. 
Some recent works proposed to relax the requirement of differential privacy to prevent membership/property attacks while provide better model utility. 
For example, some regularization techniques such as dropout are helpful to prevent the information leakage of machine learning models~\cite{wu_sgld,collabrative_leakge}. Recent studies consider 
to use Stochastic gradient Langevin dynamics (SGLD)~\cite{wu_sgld,mou_sgld}.  
In this paper, to prevent the information leakage in deep transfer learning, we employ SGLD as our optimizer to optimize deep models and show its effectiveness in reducing information leakage. 
Prior work~\cite{wu_sgld} demonstrates that SGLD is effective in preventing membership information
leakage and provides theoretical bounds for membership privacy loss. Empirically, in this paper, SGLD is also effective for preventing the leakage of the property information while provides comparable model performance compared with those non-private training methods.

To examine the effectiveness of the defense solutions, we conduct experiments under the mapping-based setting. 
Specifically, we replace the non-private optimizer (i.e., SGD) with SGLD and train the source and target models from scratch. 
The overall result is shown in Table~\ref{tab:defnese}. We observe that the use of SGLD can prevent the information leakage of the training dataset, based on the above attack metrics for evaluating the information leakage. In contrast to the original optimizer, SGLD significantly reduces the attack AUC score of the inference attack from 0.7766 to 0.6862 in Prop-sex, and 0.5885 to 0.5442 in Prop-race, 
while achieves better model utility in terms of the task AUC score especially in Prop-race (0.7750 to 0.8012). 
In both datasets, SGLD achieves the best task AUC and slightly outperforms the original task results. We infer the boost of the task AUC score is due to the regularization effect of the noises injected in SGLD, which can prevent model overfitting as pointed out in the previous works~\cite{wu_sgld,mou_sgld}. 

We also conduct experiments of DP-SGD introduced in the prior work~\cite{dp_sgd}. As the result shows, although DP-SGD can achieve better anti-attack ability than SGLD, it comes with considerably model utility decrease (the task AUC score decreases from 0.8466 to 0.7662 in Prop-sex, and 0.7750 to 0.7192 in Prop-race ). Furthermore, experiments on Dropout are also performed.
We can observe that Dropout can also prevent privacy leakage. 
We also note that when the drop ratio increases, the anti-attack ability of Dropout increases but model utility decreases drastically. 
As a conclusion, through experiments, we found that SGLD can be a good alternative to differentially private optimization methods and it can achieve a better trade-off between model utility and anti-attack ability.

\begin{table}[]
\caption{A comparison of defenses for property attack.
}
\vspace{-0.1cm}
\label{tab:defnese}
\begin{tabular}{l | c c | c c }
\toprule
  & \multicolumn{2}{c|}{Prop-sex} & \multicolumn{2}{c}{Prop-race} \\
  & AUC$_{Task}$ & AUC$_{Attack}$ & AUC$_{Task}$ & AUC$_{Attack}$ \\
\hline
Original & 0.8466 & 0.7766 & 0.7750 & 0.5885 \\\hline
SGLD & 0.8501 & 0.6862 & 0.8012 & 0.5442 \\
DP-SGD & 0.7662 & 0.6656 & 0.7192 & 0.5172 \\\hline
Dropout-0.1 & 0.7449 & 0.7126 & 0.7748 & 0.6040\\
Dropout-0.5 & 0.5881 & 0.6132 & 0.6194 & 0.5199\\
Dropout-0.9 & 0.5240 & 0.5188 & 0.5267 & 0.5041\\
\bottomrule
\end{tabular}
\vspace{-0.1cm}
\end{table}

\section{Industrial Application}
Witness the effectiveness of the SGLD, we conduct additional experiments to examine whether it is able to prevent information leakage in an industrial application.
The main task is to predict whether a user will use the coupon and the inference task is to predict the property “marital status: married” (the Pearson correlation between these labels is -0.03986). Follow the original application, here we adopt a fully-shared model \cite{liu2017adversarial} under the parameter-based setting.
As shown in Table~\ref{tab:industry}, we find the original TL method does help to improve the target domain performance, boosting AUC from 0.7513 (Target-only) to 0.7704. However it also suffers from information leakage with an attack AUC of 0.5553. By combining SGLD with transfer learning, with a minor decrease of task performance (-0.8\%), the information leakage can be significantly reduced by 3.8\%.
\begin{table}[ht!]
\vspace{-0.1cm}
\caption{A comparison of defenses for batch property attack on the ``Marketing'' dataset in terms of AUC.}
\vspace{-0.1cm}
\label{tab:industry}
\begin{tabular}{l | c | c  c| c c}
\toprule
& Target & \multicolumn{2}{c|}{Original}
& \multicolumn{2}{c}{Defense}  \\
& only & TL & Attack & TL+SGLD & Attack \\\hline
AUC & 0.7513 & 0.7704 & 0.5553 & 0.7625(-0.8\%) & 0.5176(-3.8\%)\\
\bottomrule
\end{tabular}
\vspace{-0.1cm}
\end{table}



\section{Related Work}
\label{sec:related_work}

\nosection{Deep Transfer Learning Models}
With the success of deep learning, deep transfer learning has been widely adopted in various applications~\cite{pan2009survey,tan2018survey,DBLP:conf/sp/NasrSH19}. 
According to our categorization based on the potential information leakage,  transfer learning models can be summarized into three types, i.e., model-based, mapping-based, and parameter-based,
or a hybrid of the different types\cite{long2017deep}.  

Model-based methods, such as model fine-tuning, generally first pre-train a model using the source domain data and then continue training on the target domain data\cite{he2016deep,howard2018universal,devlin2019bert}. Mapping-based methods aim to align the hidden representations by explicitly reducing the marginal/conditional distribution divergence between source and target domains which are measured by some distribution difference metrics.  Commonly used metrics include variants of Maximum Mean Discrepancy(MMD), Kullback-Leibler Divergence, Wasserstein distance, and etc~\cite{long2015learning,long2017deep,Rozantsev,tan2018survey,DBLP:conf/sp/NasrSH19}.
Parameter-based methods transfer knowledge by jointly updating a shared network to learn domain-invariant features across domains, which is mainly achieved by parameter sharing~\cite{NIPS2014_5347,liu2017adversarial,yang2017transfer}.
Further works improve parameter-based methods by incorporating adversarial training to better learn domain-invariant shared representations~\cite{liu2017adversarial}.

There are few studies~\cite{sharma2019secure,liu2018secure} for analyzing privacy leakages for general machine learning models or in a federated learning setting, however, there is no such privacy analysis work for transfer learning models. To bridge this gap, we first provide a general categorization of deep transfer learning models based on information leakage types and conduct thorough privacy analysis through building specific attacks against different transfer learning paradigms. 
We we also examine several general defense solutions to alleviate information leakage for the three paradigms. 
The privacy preserving models for transfer learning are rarely studied. Prior work~\cite{liu2018secure} proposed a secure transfer learning algorithm under the mapping-based setting, where homomorphic encryption was adopted to ensure privacy at the expenses of efficiency and some model utility loss due to the Taylor approximation.  A follow-up work in~\cite{sharma2019secure} further enhanced the security and efficient for the same problem setting by using Secret Sharing technique. A recent work by~\cite{DBLP:journals/corr/abs-1811-09491} employed a privacy preserving logistic regression with $\epsilon$-differential privacy guarantee under a hypothesis transfer learning setting.


\nosection{Membership Inference} The study~\cite{shokri2017membership} developed the first membership attack against machine learning models with only black-box access 
using a shallow training method. 
This method assumes that we have some prior knowledge of the data for training the attack classifier is from the same distribution as the original training data. 
Later study of \cite{long2017towards} followed the idea of shallow training and explored two more targeted membership attacks, i.e., frequency-based and distance-based.
The study\cite{salem2018ml} further relaxed key attack assumptions of \cite{shokri2017membership} and demonstrated more applicable attacks.
Aside from the black-box setting, these studies~\cite{melis2019exploiting,nasr2019comprehensive} examined the membership attacks against federated/collaborative learning under the white-box setting, where an adversary can access the model and potentially is able to observe/eavesdrop the intermediate computations at hidden layers. They share a similar idea that leverages the gradients or model snapshots to produce the labeled examples for training a binary membership classifier.
This work~\cite{hayes2019logan} presented the first membership attacks on both black-box and white-box for generative models, in particular generative adversarial networks~(GANs).

\nosection{Property Inference}
Property attack aims to infer the properties hold for the whole or certain subsets of the training data.
Prior works \cite{ateniese2015hacking,fredrikson2015model,ganju2018property} studied property inference attacks that characterize the entire training dataset. A property attack was developed~\cite{ganju2018property} based on the concept of permutation invariance for fully connected neural networks, with the assumption that adversary has white-box knowledge. 
Concurrently, the study~\cite{melis2019exploiting} developed attacks under the collaborative learning setting, where they focus on inferring properties for single batches of training inputs.







\section{Conclusion}
\label{sec:conclusion}
In this study, we provide a general categorization of different deep transfer learning paradigms depending on how the domains interact with each other. Based on that, we then analyze their respective privacy leakage profiles, design different attack models for each paradigm and provide potential solutions to prevent these threats. Extensive experiments have been conducted to examine the potential privacy leakage and effectiveness of defense solutions.

\bibliographystyle{ACM-Reference-Format}
\bibliography{sample-base}

\end{document}